\documentclass[letterpaper, 10 pt, conference]{ieeeconf}  

\IEEEoverridecommandlockouts                              

\usepackage{graphicx}

\usepackage{tabularx}
\usepackage{booktabs}
\usepackage{bm}
\usepackage{caption}

\usepackage[table, dvipsnames]{xcolor}

\usepackage{amsmath,amssymb}

\title{\LARGE \bf 3DMODT: Attention-Guided Affinities for Joint Detection \& Tracking in 3D Point Clouds}

\author{Jyoti Kini$^{1}$, Ajmal Mian$^{2}$ and Mubarak Shah$^{1}$ 
\thanks{$^{1}$Center for Research in Computer Vision, University of Central Florida, USA 
        {\tt\small jyoti.kini@knights.ucf.edu, shah@crcv.ucf.edu}}%
\thanks{$^{2}$University of Western Australia
        {\tt\small ajmal.mian@uwa.edu.au}}%
}

\begin{document}

\maketitle
\thispagestyle{empty}
\pagestyle{empty}

\begin{abstract}

We propose a method for joint detection and tracking of multiple objects in 3D point clouds, a task conventionally treated as a two-step process comprising object detection followed by data association. Our method embeds both steps into a single end-to-end trainable network eliminating the dependency on external object detectors. Our model exploits temporal information employing multiple frames to detect objects and track them in a single network, thereby making it a utilitarian formulation for real-world scenarios. Computing affinity matrix by employing features similarity across consecutive point cloud scans forms an integral part of visual tracking. We propose an attention-based refinement module to refine the affinity matrix by suppressing erroneous correspondences. The module is designed to capture the global context in affinity matrix by employing self-attention within each affinity matrix and cross-attention across a pair of affinity matrices. Unlike competing approaches, our network does not require complex post-processing algorithms, and processes raw LiDAR frames to directly output tracking results. We demonstrate the effectiveness of our method on the three tracking benchmarks: JRDB, Waymo, and KITTI. Experimental evaluations indicate the ability of our model to generalize well across datasets.
\end{abstract}
\section{INTRODUCTION}

Rapid advancements in 3D sensing technology and the recent commercial interest in autonomous driving \cite{kiran2021deep, weng2020sequential} and assistive robots \cite{manglik2019forecasting, sun2020we}, have sparked great research interest in multiple object tracking (MOT) in 3D point clouds obtained from LiDAR (Light Detection and Ranging) sensors.  MOT aims to predict the trajectories of distinct objects in a given sequence. In 2D tracking, two major paradigms: (a) tracking-by-detection \cite{shenoi2020jrmot, zhang2019robust, wojke2017simple}, and (b) joint detection and tracking \cite{huang2021joint, peng2020chained, lu2020retinatrack} have commonly been used to advance research in this direction. Although numerous 2D MOT approaches \cite{wu2021track, lu2020retinatrack, peng2020chained, zeng2021motr, mouawad2022fastervideo, wang2021joint, tapu2017deep} have leveraged the benefits of jointly performing detection and tracking in a single unified framework, several open challenges have afflicted this long-standing research problem of MOT in 3D point clouds.

\begin{figure}[t]
\centering
\includegraphics[width=\columnwidth]{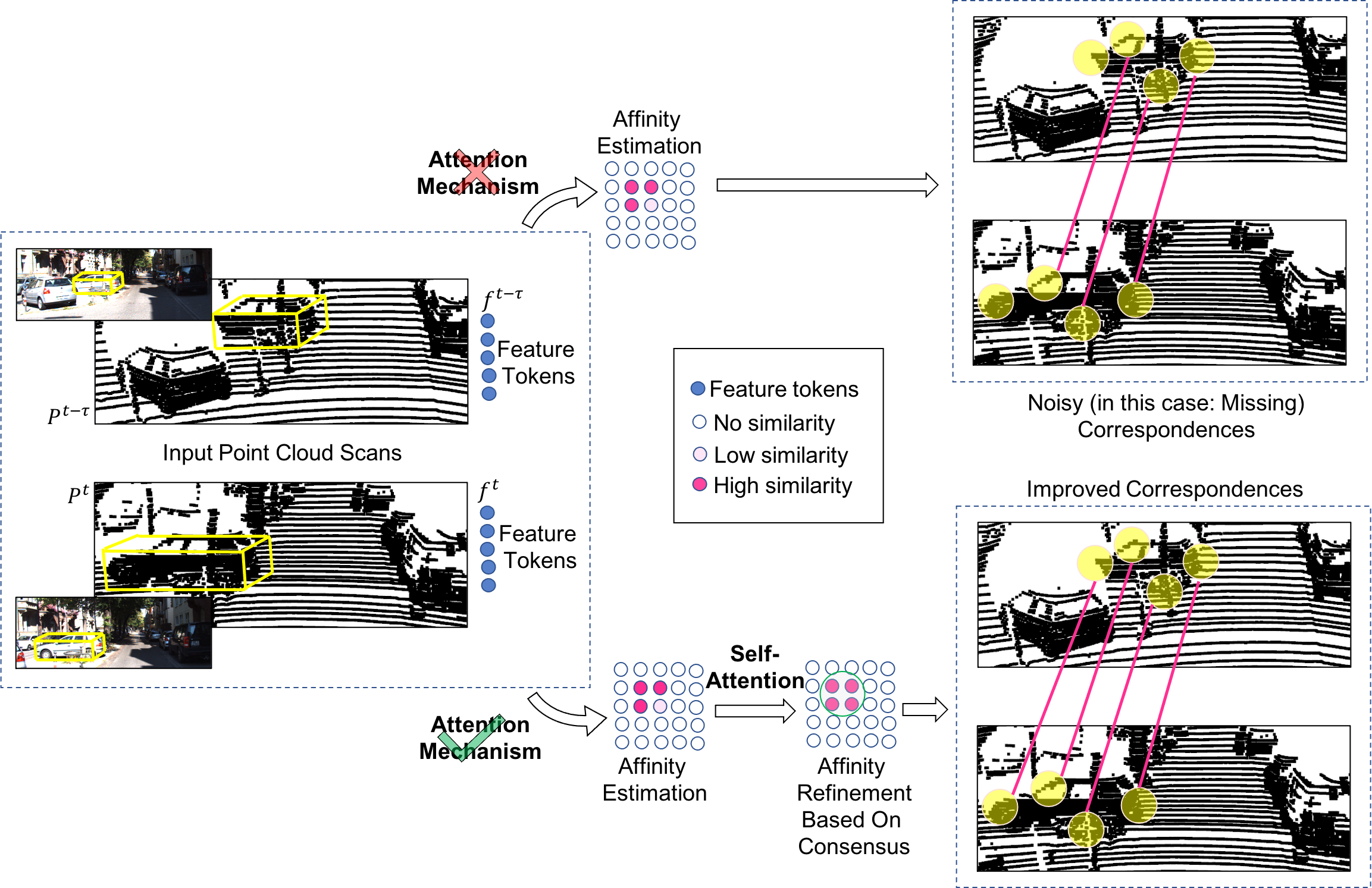}
\caption{\textbf{Overview of our attention-guided affinity refinement.}
Feature tokens of two consecutive point cloud scans are used to construct an affinity matrix that stores dense similarities between the tokens. We use attention (bottom) to refine the affinities. Initially, the \textit{light pink} node is linked incorrectly. 
However, self-attention on affinities seeks consensus from the other nodes (i.e. \textit{dark pink} nodes). Supporting cues provided by the other nodes boost the correct correspondences and weaken the noisy correspondence. The input to our network is point cloud scans. RGB images and bounding boxes are shown for visualization purposes only.}
\label{fig:Teaser}
\end{figure}

3D point cloud-based tracking provides robust depth perception, motion prediction, and path planning for critical robot navigation tasks. However, LiDAR point clouds are characterized by challenges such as data sparsity, uneven sampling density, lack of texture, etc. To accomplish tracking across point cloud scans, typically object features are used to compute robust affinities for comprehensive data association (correspondences between the detected objects across multiple scans). Popular 3D MOT approaches \cite{aakash2021pcdan, Weng2020_AB3DMOT} rely heavily on external off-the-shelf detectors to localize objects of interest. This not only adds an overhead to the tracking, but sub-par detections also inherently affect the overall performance of the tracker. Moreover, given the sparse and non-uniform nature of the point clouds, simple affinities estimation in current 3D joint detection and tracking approach \cite{luo2021exploring} leads to noise in correspondences, which are crucial for robust tracking.

We propose an end-to-end learning framework for joint detection and tracking in 3D point clouds. As opposed to the conventional use of attention on features, we take a unique approach and propose self and cross-attention on the affinity matrices. Three consecutive point cloud frames are passed through a single end-to-end network to extract features employing transformer encoder, followed by the computation of pairwise affinity matrices. The affinity matrices are refined (see Figure \ref{fig:Teaser}) using an attention mechanism and intermediate supervision. Thereafter, 3D tracking offsets are computed. The final output is supervised using ground truth object detections and is composed of regressed values of object centers and bounding boxes in 3D. The output object detection information and tracking offsets are collectively used to generate tracks. Experiments on the JRDB, Waymo and KITTI 3D tracking benchmarks show that our end-to-end network achieves competitive results w.r.t. existing MOT methods, {\em without} the need for an external object detector during inference or extensive post-processing of the tracking results. To summarize our contributions:
\begin{itemize}
\item{We propose a novel 3D point cloud-based joint detection and tracking method that eliminates the reliance on external object detector.} 
\item{We develop an end-to-end network that directly processes raw point clouds to produce multiple object tracks, without the need to pre-process the point clouds to eliminate detection uncertainties or post-process the output results using extensive association algorithms.}
\item{We design a novel attention-based refinement module to capture global context of affinities, which is essential to refining the affinity matrices capturing correspondences across the point cloud scans and thereby improving tracking accuracy.}
\item{We show competitive performance on the KITTI, JRDB, and Waymo tracking benchmarks as well as the generalization ability of our method across datasets.}
\end{itemize}

\begin{figure*}[t]
\centering
\includegraphics[width=\textwidth]{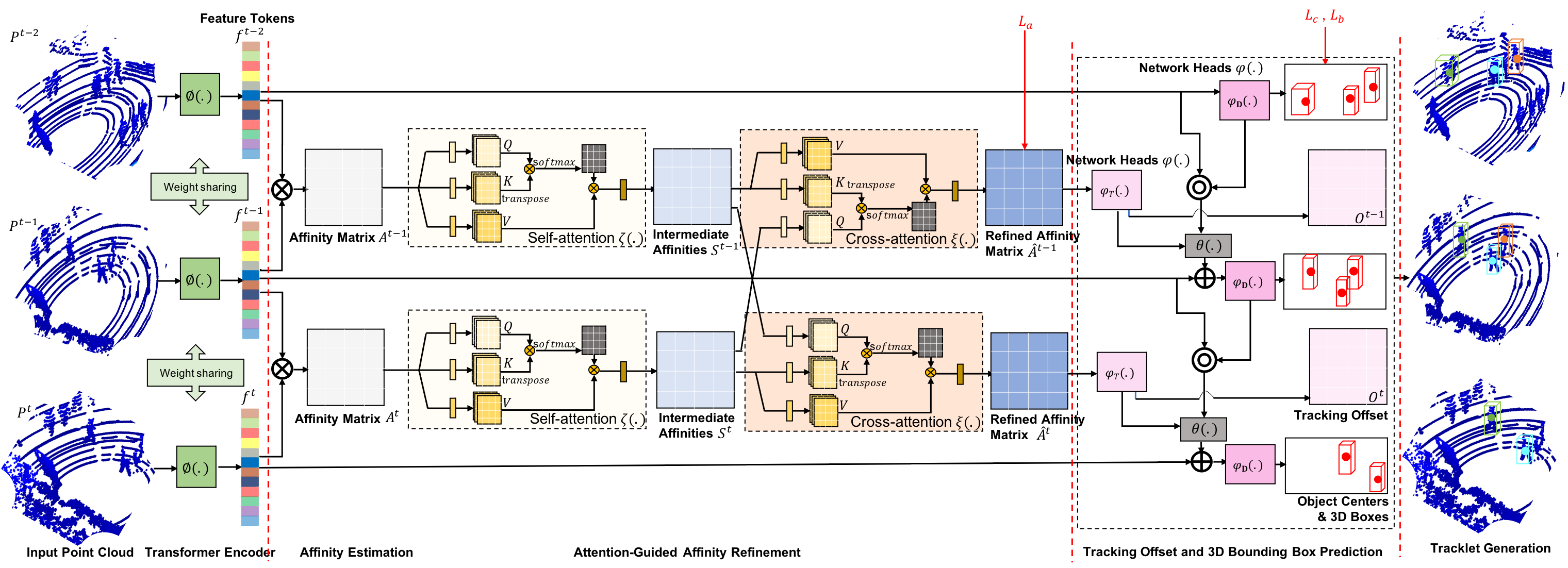}
\caption{\textbf{Details of the 3DMODT network.} 3DMODT performs joint detection and tracking in 3D point cloud data. Input point cloud scans $\mathcal{P} = \{P^{t}$, $P^{t-1}$, $P^{t-2}\}$ are passed through a transformer encoder $\phi(.)$. The feature tokens are used to construct affinity matrices $\mathcal{A}=\{A^t, A^{t-1}\}$
that store dense similarities between tokens in consecutive scans. Attention is then applied to the affinity matrices for refinement. The network heads $\varphi(.)$ generate (a) tracking offsets that store spatio-temporal displacements of points between consecutive scans, and (b) 3D detections. The predictions from the network heads are collectively used to generate tracklets. $L_a$, $L_c$, $L_b$ represent Forward Association Loss on Affinities, $\ell_1$ Center Offset Loss and $\ell_1$ Dimension Size Loss. $\otimes$, $\oplus$, $\odot$ denote matrix multiplication, element-wise sum and Hadamard product.} 
\label{fig:Architecture}
\end{figure*}

\section{Related Work}
\noindent
\textbf{2D Multi-Object Tracking:}
Given the continued advancements in object detection techniques \cite{ren2015faster, ren2020ufo, ren2020instance}, most 2D multi-object tracking methods use the tracking-by-detection approach. This pipeline involves first detecting the objects and then associating the detections across frames. The association step can be categorized into two groups: appearance-based and motion-based. Appearance-based methods \cite{zou2020joint, zheng2019joint} establish object correspondences across frames by using re-identification features. On the other hand, motion-based approaches use temporal modeling to update the object detections using distance or intersection-over-union (IOU) metric. State estimation using the Kalman filter \cite{kalman1960new} is a common practice amongst these approaches. Motion modeling \cite{peng2020chained, zhou2020tracking} by predicting motion offsets is another typical strategy. Despite the popularity of the tracking-by-detection paradigm, joint detection and multi-object tracking has gained momentum due to the missing connection between detections and associations in the former case.
\\
Joint detection and tracking methods co-optimize both tasks, thereby boosting performance gains. FairMOT \cite{zhang2021fairmot} and JDE \cite{wang2020towards} introduce a re-identification module into the detector, while GSDT \cite{wang2021joint} incorporates Graph Neural Networks to accomplish the task. Inspired by the success of Transformers \cite{arnab2021vivit}, TrackFormer \cite{meinhardt2022trackformer} and TransTrack \cite{sun2020transtrack}  employ query-key mechanisms to follow object locations.
\\
\noindent
\textbf{3D Multi-Object Tracking:}
Although 3D information is crucial for depth perception, the data is inherently challenging to work with. Therefore, the research trend in 3D space ensues the progress in 2D space. Similar to its 2D counterpart, tracking-by-detection is a more commonly used practice in 3D multi-object tracking. The availability of sophisticated detectors has facilitated the growth in this space. VoxelNet \cite{zhou2018voxelnet}, SECOND \cite{yan2018second}, PointPillar \cite{lang2019pointpillars} are LiDAR-only detectors.  In the tracking-by-detection approach, detections are followed by association. AB3DMOT \cite{Weng2020_AB3DMOT} uses the Kalman filter for state motion estimation. PnPNet \cite{liang2020pnpnet} learns affinity matrix using 3D features and trajectories. CenterPoint \cite{yin2021center} incorporates a regression head to estimate velocity. Chiu et al. \cite{chiu2021probabilistic} enhances association by introducing distance metric. However, in LiDAR point cloud scans, object associations rely solely on motion modeling and lack appearance cues. Therefore, most tracking approaches, additionally, employ complex association algorithms (e.g. Hungarian algorithm) for bipartite matching. Some other detectors, namely ContFuse \cite{liang2018deep}, MMF \cite{liang2019multi}, CLOCS \cite{pang2020clocs}, 3D-CVF \cite{yoo20203d}, fuse camera images with LiDAR point clouds to introduce appearance information. However, fusing camera data introduces sensor calibration overhead. 
\\
Similar to 2D space, 3D joint detection and tracking attempts to leverage the power of optimizing the two tasks in a single end-to-end network. JRMOT \cite{shenoi2020jrmot} combines re-identification, detection, and tracking in a single framework. Similar to our work, SimTrack \cite{luo2021exploring} works solely with LiDAR data for joint detection and tracking by introducing a motion estimation branch to eliminate the heuristic matching step. In our work, however, we focus on refining the affinity matrix, which maps the correspondences across point cloud scans, using the attention-based refinement module.
\\
\noindent
\textbf{Attention Mechanism:}
Over the past few years, the attention mechanism has gained accolades due to its focus on non-local operations \cite{wang2018non}. Its variants (self-attention and cross-attention) have been used in the spatial and temporal domains for dynamic information aggregation \cite{gao2022aiatrack}. We use the concept of the global receptive field provided by the attention mechanism to refine the affinity matrix. We suppress the erroneous correspondences in the affinity matrix by seeking consensus among the affinities using attention.

\section{Approach}
Figure \ref{fig:Architecture} provides detailed schematics of the proposed 3DMODT, which has four main  building blocks: (1) transformer encoder for feature extraction, (2) affinity computation and attention-based refinement, (3) tracking offset and 3D detection prediction, and (4) data association and tracklet-generation. We extract feature tokens $f^t$, $f^{t-\tau}$ from consecutive point cloud scans $P^{t}, P^{t-\tau}$. Thereafter, we use these feature tokens to construct affinity matrix ${A}^{t}$ that stores dense similarity matches of tokens associated with $t$ and $t-\tau$. The affinity matrix is then passed through attention mechanism leveraging global receptive fields to get refined affinity matrix $\hat{A}^{t}$. Network heads are employed to compute: (a) tracking offsets $O^t_{i, j, k}$ that store spatio-temporal displacements for all points, denoted by 3D locations (i, j, k), from time $t$ to the corresponding points at $t-\tau$, and (b) regress 3D object information (center, bounding box, rotation). 3D bounding box information combined with tracking offset is used to generate tracklets. Our model processes three point cloud scans simultaneously in a single pass to leverage self-attention and cross-attention across affinity matrices, thereby improving the temporal context. The four building blocks of our model are explained below.

\subsection{Feature Extraction}
\label{Features}
The input to our network is three point cloud scans $\mathcal{P} = \{P^{t}, P^{t-1}, P^{t-2}\} \in \mathbb{R}^{N\times3}$ captured at time step $t, t-1, t-2$ and consisting $N$ 3D points. Each point cloud is passed through a transformer encoder (feature extractor) $\phi(.)$ \cite{Zhao_2021_ICCV} to generate feature tokens $\hat{\mathcal{P}} = \{f^{t}, f^{t-1}, f^{t-2}\} \in \mathbb{R}^{M\times dim}$. Here, $M \subset N$ and $dim$ corresponds to 64 feature representation channels + 3 channels of positional information.  

\subsection{Attention-Guided Affinity Refinement}
\label{Affinity}
\noindent
\textbf{Affinity Estimation:} We process the feature tokens to construct affinity matrices. These affinity matrices store similarity scores between tokens at location index $d$ for a given scan and tokens at location index $e$ for its successive scan, with indexes $d = 1,....,M$ and $e = 1,....,M$. Given the feature tokens $f^t_d$ (or simply $f^t$) and  $f^{t-\tau}$ (or simply $f^{t-\tau}_e$), cosine similarity is applied to compute the affinity matrix. The affinity matrix $A^{t}_{d, e}$ (or simply $A^t$) across time $t$ and ${t-\tau}$ is given by:
\begin{align}
    A^{t}_{d, e} = \frac{f^{t}_{d} \cdot f^{t-\tau}_{e}}{||f^{t}_{d}|| \: ||f^{t-\tau}_{e}||}
\end{align}
Our network computes two affinity matrices $\mathcal{A} = \{A^t, A^{t-1}\}$ across time steps ($t$, $t-1$), and ($t-1$, $t-2$) respectively. 
The computed affinity matrices $\mathcal{A}$ exhibit noisy affinities due to inter-object similarities and intra-object variations caused by viewpoint changes and occlusions. Therefore, we apply attention mechanism to suppress the erroneous similarities by capturing global context. Typically, attention is applied to improve input feature representation. In our case, however, we take a novel approach and introduce attention on affinity matrices, as opposed to the input feature embeddings.
\\
\noindent
\textbf{Affinity Refinement:}
Attention-guided affinity refinement is a two-step process. First, we apply self-attention on the individual affinity matrix $A^t$ and $A^{t-1}$ to get intermediate affinity matrix $S^t$ and $S^{t-1}$ respectively. This is followed by cross-attention applied on the intermediate affinity matrices $\mathcal{S} = \{S^t, S^{t-1}\}$ to yield the refined affinity matrices $\hat{\mathcal{A}} = \{\hat{A}^t, \hat{A}^{t-1}\}$.

\noindent\textit{\underline{Self-attention}}, applied on individual affinity matrix, seeks consensus from the similarities within the affinity matrix and suppresses erroneous correspondences. We employ convolutions followed by projection layer \cite{zhou2022pttr} to transform the affinity matrices into query $Q$, and set of key-value ($K$, $V$) pair vectors. Thereafter, scalar product computed between $Q$ and $K$ is passed through a softmax activation, and the  resultant output is used to re-weight $V$. The self-attention on individual affinity matrix $\zeta(.)$ is formulated as:
\begin{align}
    \zeta(A^t) = \text{softmax}({\bar{Q}^t\bar{K}^t}^{\mathsf{T}})\bar{V}^t,
\end{align}
\begin{align}
    \zeta(A^{t-1}) = \text{softmax}({\bar{Q}^{t-1}\bar{K}^{t-1}}^{\mathsf{T}})\bar{V}^{t-1},
\end{align}
where $\bar{Q} = QW_q$, $\bar{K} = KW_k$, $\bar{V} = VW_v$ represent various linear transformations \cite{zhou2022pttr}, and $W_q, W_k, W_v$  denote transformation weights for query, key and value. 

\noindent\textit{\underline{Cross-attention}}, on the other hand, is applied across affinity matrices. It aims to suppress erroneous correspondences in a given affinity matrix, by seeking agreement between similarities of the given affinity matrix and its successive affinity matrix. Applying cross-attention on affinity matrices implicitly increases the temporal context. The cross-attention across affinity matrices $\xi(.)$ is given by: 
\begin{align}
    \xi(A^t) = \text{softmax}({\bar{Q}^{t-1}\bar{K}^t}^{\mathsf{T}})\bar{V}^t,
\end{align}
\begin{align}
    \xi(A^{t-1}) = \text{softmax}({\bar{Q}^{t}\bar{K}^{t-1}}^{\mathsf{T}})\bar{V}^{t-1}.
\end{align}

\subsection{Tracking Offset and 3D Bounding Box Prediction}
\label{Predictions}
The initial feature tokens $\hat{\mathcal{P}}$ and refined affinity matrices $\hat{\mathcal{A}}$ are processed through network heads $\varphi(.)_T$ and $\varphi(.)_D$\ to generate tracking offsets and 3D detections respectively. 
We generate tracking offset $O^{t}_{i, j, k} \in \mathbb{R}^{N\times3}$ that estimates the spatio-temporal displacement for point $(i, j, k)$ at time $t$ to its corresponding point at time $t-\tau$. 
In addition, several light-weight convolutions 
are used to predict 3D bounding boxes $\mathcal{B} = \{B_k\}$. Each bounding box $B = (c_x, c_y, c_z, b_w, b_l, b_h, \alpha)$ comprises object center $(c_x, c_y, c_z)$, object size $(b_w, b_l, b_h)$ and rotation in yaw $\alpha$. The values are regressed in logarithmic scale. 

\subsection{Tracklet Generation}
\label{Tracklet}
To generate tracklets, we connect the 3D detection bounding boxes $\mathcal{B}$ across time. For time step $t$, we first look for bounding box associations within a limited space in $t-1$, as prompted by the tracking offsets. In case of unmatched bounding boxes, we increase the search space in $t-1$. Basically, given time step $t$, we use the bounding box center location $(c_x, c_y, c_z)$, add it to the tracking offset $O^{t}_{i, j, k}$ for that location and associate the closest matching bounding box within the area at time $t-1$. If the bounding box does not find a match, we compute cosine similarity of its embedding with all unmatched embeddings at time $t-1$, and associate based on highest similarity. In case the bounding box still fails to find a match, we initiate a new tracklet.


\setlength{\tabcolsep}{2.8pt}
\begin{table*}[!t]
\scriptsize
\begin{minipage}[b]{0.56\linewidth}
    \begin{center}
        \renewcommand{\arraystretch}{1}
        \caption{\textbf{Results on JRDB dataset.} Our method exhibits competitive results on the JRDB test set without using additional input modality, any external detector, or a complex post-processing algorithm. Here, $L$ implies LiDAR data, and $I$ refers to camera feed.}
        \label{table:JRDB}
        \begin{tabularx}{8.2cm}{@{}l*{4}c*{2}r@{}}
        \toprule
        \rowcolor{gray!20}{Method} & 
        {Input} &
        {External} &
        {Complex} &
        {\emph{MOTA}}$\uparrow$ &  
        {\emph{MOTP}}$\uparrow$ & 
        {IDS}$\downarrow$ \\
        \rowcolor{gray!20}&{Data}&{Detector}&{Post-Processing}&&&\\
        \midrule
        AB3DMOT \cite{Weng2020_AB3DMOT} & $L$ & $\checkmark$ & $\checkmark$ & 19.34 & 42.01 & 6,179 \\
        JRMOT \cite{shenoi2020jrmot} & $L + I$ & $\times$ & $\checkmark$ & 20.15 & 42.44 & 4,216 \\
        PC-DAN \cite{aakash2021pcdan} & $L$ & $\checkmark$ & $\checkmark$ & \textbf{22.56} & 6.01 & 26,022 \\
        \midrule
        3DMODT & $L$ & $\times$ & $\times$ & 21.74 & \textbf{47.61} & \textbf{3,805} \\
        \bottomrule
    \end{tabularx}
    \end{center}
\end{minipage}\hspace{10pt}
\begin{minipage}[b]{.40\linewidth}
\begin{center}
        \renewcommand{\arraystretch}{1}
        \caption{\textbf{Generalization performance.} Demonstration of the ability of the network to generalize across datasets.} 
        \label{table:Generalisation}
        \begin{tabularx}{3.7cm}{@{}l*{2}{c}c@{}}
        \toprule
        \rowcolor{gray!20}{Training Set} & 
        {Testing Set} & 
        {\emph{MOTA}}$\uparrow$ \\
        \midrule
        JRDB & JRDB & 21.74\\
        Waymo & JRDB & 17.83\\
        KITTI & KITTI & 92.33 \\
        Waymo & KITTI & 83.48\\
        \bottomrule
    \end{tabularx}
    \end{center}
\end{minipage}
\end{table*}

\setlength{\tabcolsep}{4pt}
\begin{table*}[!t]
    \begin{center}
    \renewcommand{\arraystretch}{1}
        \caption{\textbf{Results on Waymo Open dataset.} We report the best tracking results on the Waymo vehicle validation set and the values are in the format LEVEL\_1 \// LEVEL\_2. Here, $L$ implies LiDAR data, and $I$ refers to camera feed.}
        \label{table:Waymo}
        \begin{tabularx}{14.1cm}{@{}l*{8}{c}c@{}}
        \toprule
        \rowcolor{gray!20}{Method} &
        {Input} &
        {External} &
        {Complex} &
        {\emph{MOTA}}$\uparrow$ &
        {\emph{MOTP}}$\uparrow$ &
        Miss$\downarrow$ & 
        Miss Match$\downarrow$ &
        {\emph{FP}}$\downarrow$ \\
        \rowcolor{gray!20}&{Data}&{Detector}&{Post-Processing}&&&&&\\
        \midrule
        Sun et al. \cite{sun2020scalability} & $L$ & $\checkmark$ & $\checkmark$ & 42.5 \// 40.1 &  18.6 \// 18.6 & 40.0 \// 43.4  & 0.14 \// 0.13 & 17.3 \// 16.4 \\
        CenterPoint \cite{yin2021center} & $L$ & $\times$ & $\times$ & 51.4 \// 47.9 & 17.6 \// 17.6 &  47.7 \// 41.4  & 0.19 \// 0.18 & 10.7 \// 10.6 \\
        SimTrack \cite{luo2021exploring} & $L$ & $\times$ & $\times$ & 53.1 \// 49.6 & 17.4 \// 17.4 & 35.5 \// 39.8 & 0.20 \// 0.19 & 11.2 \// 10.5\\ 
        \midrule
        3DMODT & $L$ & $\times$ & $\times$ & \textbf{55.9} \// \textbf{51.2} & \textbf{18.9} \// \textbf{18.9} & \textbf{33.1} \// \textbf{36.7} & \textbf{0.13} \// \textbf{0.12} & \textbf{10.3} \// \textbf{10.2} \\ 
        \bottomrule
    \end{tabularx}
    \end{center}
\end{table*}

\setlength{\tabcolsep}{3.8pt}
\begin{table*}[t]
    \begin{center}
        \renewcommand{\arraystretch}{1}
        \caption{\textbf{Results on KITTI tracking benchmark.} Our tracker demonstrates the best 3D tracking performance on the KITTI cars validation set. Here, $L$ implies LiDAR data, and $I$ refers to camera feed.}
        \label{table:KITTI}
        \begin{tabularx}{13.1cm}{@{}l*{9}{c}c@{}}
        \toprule
        \rowcolor{gray!20}{Method} & 
        Input &
        External &
        Complex &
        {\emph{sAMOTA}}$\uparrow$ &
        {\emph{AMOTA}}$\uparrow$ &
        {\emph{AMOTP}}$\uparrow$ &
        {\emph{MOTA}}$\uparrow$ &
        {\emph{MOTP}}$\uparrow$\\
        \rowcolor{gray!20}&{Data}&{Detector}&{Post-Processing}&&&&&\\
        \midrule
        mmMOT \cite{zhang2019robust} & $L + I$ & $\checkmark$ & $\checkmark$ & 70.61 & 33.08 & 72.45 & 74.07 & 78.16 \\
        FANTrack \cite{baser2019fantrack} & $L + I$ & $\checkmark$ & $\times$ & 82.97 & 40.03 & 75.01 & 74.30 & 75.24\\
        AB3DMOT \cite{Weng2020_AB3DMOT} & $L$ & $\checkmark$ & $\checkmark$ & 93.28 & 45.43 & 77.41 & 86.24 & 78.43\\
        PC-TCNN \cite{wu2021tracklet} & $L$ & $\times$ & $\times$ & 95.44 & 47.64 & $-$ & $-$ & $-$ \\
        3D DetecTrack \cite{koh2022joint} & $L + I$ & $\times$ & $\times$ & 96.49 & 48.87 & 81.56 & 91.46 & 82.24 \\
        \midrule
        3DMODT & $L$ & $\times$ & $\times$ & \textbf{96.91} & \textbf{49.01} & \textbf{82.14} & \textbf{92.33} & \textbf{83.52}\\
        \bottomrule
    \end{tabularx}
    \end{center}
\end{table*}

\setlength{\tabcolsep}{3pt}
\begin{table}[!t]
    \begin{center}
        \renewcommand{\arraystretch}{1}
        \caption{\textbf{Ablation  on JRDB validation set.} Components are added sequentially to highlight their contribution.} 
        \label{table:Ablation}
        \begin{tabularx}{4.2cm}{@{}l*{3}{c}c@{}}
        \toprule
        \rowcolor{gray!20}{Method} & 
        {\emph{MOTA}}$\uparrow$ &
        {IDS}$\downarrow$ \\
        \midrule
        Baseline & 77.31 & 14513\\
        + Self-Attention & 91.34 & 3562\\
        + Cross-Attention & 96.23 & 1960 \\
        \bottomrule
    \end{tabularx}
    \end{center}
\end{table}


\subsection{Proposed Loss Formulation}
We use three different losses to train 3DMODT. The first one, Forward Association Loss, is used to learn accurate correspondences in the affinity matrix computed across time $t$ and $t-\tau$. It is given by:
\begin{align}
    L_{a} = \frac{\sum({G} \odot (- \log{\hat{A}^{t}}))}{{\sum{G}}}
\end{align}
where G refers to the ground-truth affinities and $\hat{A}^{t}$ denotes the refined affinity matrix prediction.
\\
The second loss is an 
$\ell_1$ Center Offset Loss for supervision of 3D object centers only i.e.,
\begin{align}
    L_{\rm c} = \frac{1}{R}\sum_{r=1}^{R}
    \left|{\hat{C_r} - {C_r}}\right|, 
\end{align}
where $R$ is the number of 3D objects, $\hat{C} \in \mathbb{R}^{3}$ implies the predicted object center and $C$ represents the ground truth center. The third loss is an $\ell_1$ Dimension Size Loss to regress the bounding boxes, 
\begin{align}
    L_{\rm b} = \frac{1}{R}\sum_{r=1}^{R}
    \left|\hat{\gamma}_{r} - \gamma_{r}\right|, 
\end{align}
where $\gamma$ refers to the width, length and height of the 3D bounding boxes. 
The combined loss L is the weighted sum of the three losses
\begin{align}
    L =  L_{a} +  \lambda_{\rm c}L_{\rm c} +  \lambda_{\rm b}L_{\rm b},
\end{align}
where $\lambda_{\rm c}$ and $\lambda_{\rm b}$ are the loss hyper-parameters.

\section{Experiments}
We performed experiments on the JRDB dataset \cite{martin2019jrdb}, Waymo Open dataset \cite{sun2020scalability} and the KITTI tracking benchmark \cite{Geiger2012CVPR}. These datasets provide 3D point clouds captured using distinct LiDAR sensors. 

\noindent\textbf{Datasets:}
\textit{\underline{JRDB}} dataset was collected in a university campus setting and contains $360^{\circ}$ views of both indoor buildings and outdoor pedestrian spaces. It captures entities of pedestrian class in densely populated areas such as cafeterias, and classrooms, which makes it a challenging dataset. It provides 54 sequences divided into 27 training and 27 test sequences. We use the recommended 7 sequences from the training split as the validation set for our experiments.
\textit{\underline{Waymo}} dataset was captured in multiple cities, namely  San Francisco, Phoenix, and Mountain View, providing large geographical coverage compared to other autonomous driving datasets. It contains vehicle, pedestrian, cyclist, and road-sign information for tracking task. The dataset comprises 798 training, 202 validation, and 150 test sequences, each spanning 20 seconds in duration. To evaluate the generalizability of the models on previously unseen locations, the test set includes geographically held-out areas. The evaluation toolkit provides two levels of difficulty: LEVEL\_1 for assessing objects with more than 5 LiDAR points, and LEVEL\_2 for objects with at least one LiDAR point.
\textit{\underline{KITTI}} dataset captured in Germany provides car and pedestrian information for tracking benchmarks. It comprises 21 training sequences and 29 test sequences. The training sequences are further split into a training set comprising 10 sequences (3975 frames) and a validation set comprising 11 sequences (3945 frames). The original KITTI benchmark provides test data only for 2D evaluations. Weng et al. \cite{weng20203d} procedure is used to conduct 3D MOT evaluation.
\\
\noindent\textbf{Evaluation Metrics:}
We use standard 3D MOT evaluation measures \cite{weng20203d}: \textit{Multiple-Object Tracking Accuracy (MOTA)}, \textit{Average MOTA (AMOTA)}, \textit{scaled AMOTA (sAMOTA)}, degree of overlap between actual and predicted objects \textit{Multiple-Object Tracking Precision (MOTP)}, \textit{Average MOTP (AMOTP)}, \textit{False Negatives (FN)}, \textit{False Positives (FP)}, \textit{Identity Switches (IDS)}, percentage of objects tracked for atleast 80\% of their total trajectory lengths \textit{Mostly Tracked Trajectories (MT)}, percentage of objects recovered for less than 20\% of their life span \textit{Mostly Lost Trajectories (ML)}, trajectory \textit{Fragmentation (FRAG)}. 
\\
\noindent\textbf{Training Details:}
Our model exhibits a transformer encoder $\phi(.)$ \cite{guo2021pct} for feature extraction and point convolutions $\theta(.)$ \cite{thomas2019kpconv} at the output layers. We use a single Nvidia V100 GPU for training our model. For the JRDB dataset, we train the model for  40 epochs using Adam optimizer \cite{kingad2015methodforstochasticoptimization} with an initial learning rate of 0.001 and reduce it by a factor of 5 every 10 epochs. In the case of the Waymo dataset, we train the model for 30 epochs with the same initial learning rate, reduced after every 10 epochs. For the KITTI dataset, we train the model of 60 epochs using the same initial learning rate, reduced every 15 epochs.

\begin{figure*}[t]
\centering
\includegraphics[width=0.6\textwidth]{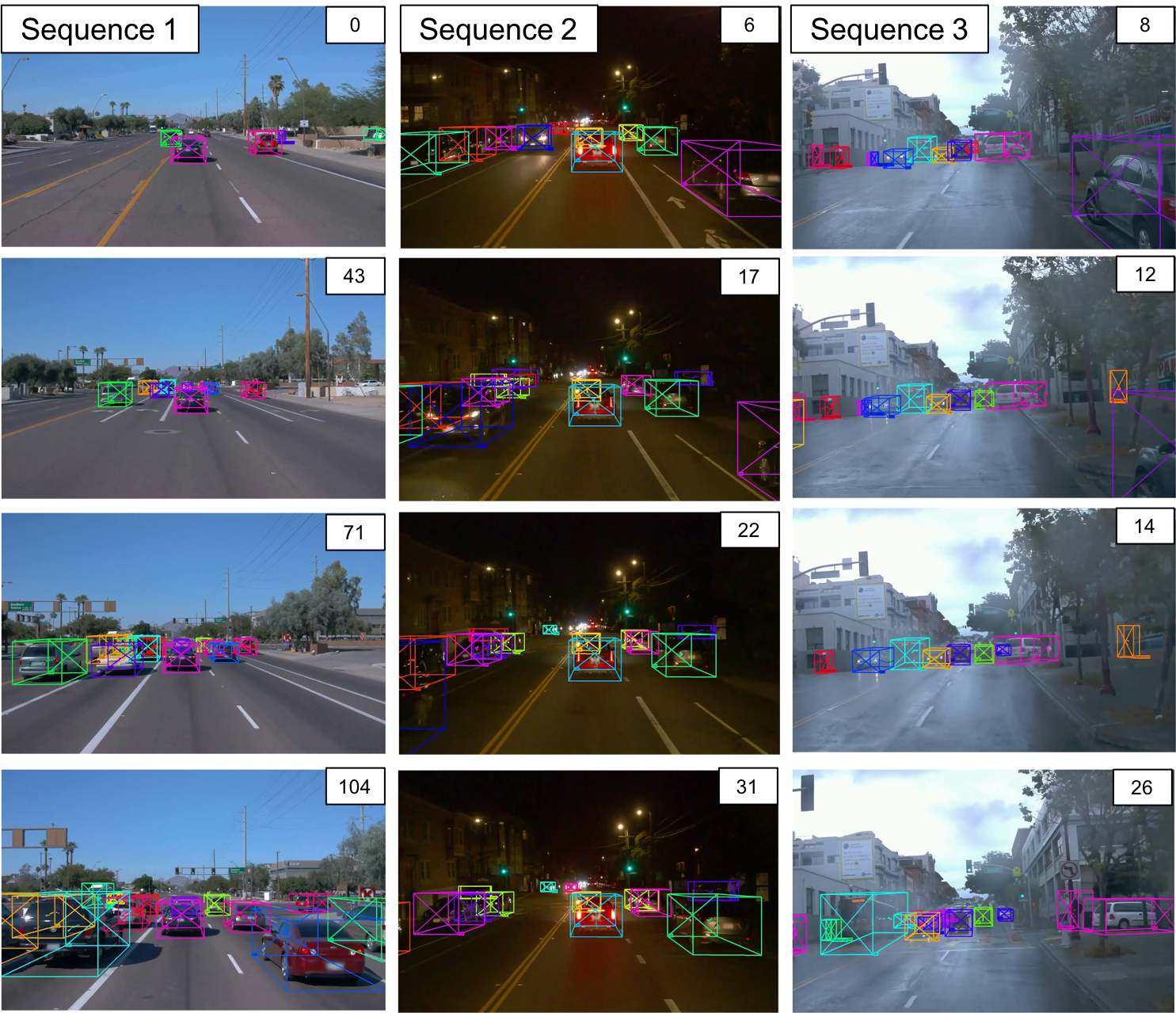}
\caption{Qualitative results of our model on diverse video sequences from Waymo dataset. Sequence 1 has long tracks. Sequence 2 has poor lighting conditions. Sequence 3 has adverse weather. 
Tracking is performed using LiDAR point clouds. 
The RGB images are for illustration only.}
\label{fig:Qualitative}
\end{figure*}

\subsection{Results and Analysis}
Table \ref{table:JRDB} presents results on the JRDB test set. We observe that the proposed method achieves a competitive MOTA of 21.74 without the complexity of an external detector or extensive post-processing steps that other methods use. Similarly, the results on the Waymo dataset in Table \ref{table:Waymo} further affirm our network's ability to generate effective tracking results.
Table \ref{table:KITTI} shows results on the KITTI car tracking benchmark. 
Unlike some of the existing approaches, our model is self contained in that it relies on the detection and motion cues (tracking offsets) generated within our network to track the objects. Moreover, we do not use additional modalities, thereby eliminating any overhead related to sensor calibrations. In Figure \ref{fig:Qualitative}, we visualize the results on Waymo. Qualitative results indicate the ability of our network to perform well under varying tracking conditions, such as long sequences, and adverse environmental situations. 

Next, we examine the generalization ability of our method across datasets. We train our model on the Waymo dataset and use JRDB and KITTI sequences for testing. In Table \ref{table:Generalisation}, we see that our model achieves good tracking results in both cross dataset settings. Our method eliminates the need for an external detector by treating detection and tracking as joint tasks in point cloud-based MOT. Given this capability and its generalizability, our model can be effectively deployed in real-world applications.

\begin{figure}
\begin{center}
    \includegraphics[width=\columnwidth]{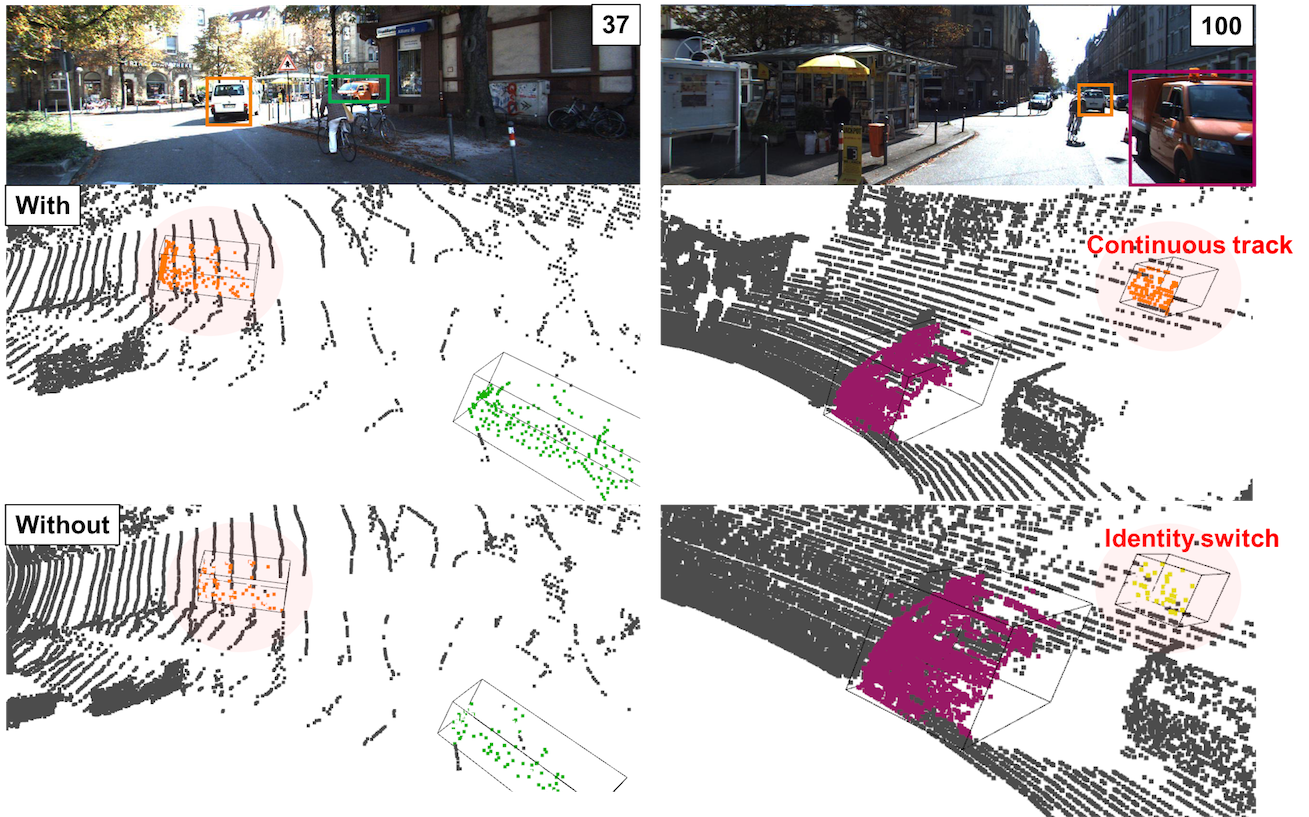}
    \caption{Qualitative results illustrating the capability of attention-guided affinity module to boost the tracking performance. Row 1 depicts RGB sequence for illustration purposes where frame numbers are shown on the top right corners. Row 2 demonstrates ability of the network (with attention-guided affinities) to track the object in orange across time steps. Row 3 exhibits ID-switch encountered on the same object in the absence of attention-on-affinities.}
    \label{fig:Ablation_Tracking}
\end{center}
\end{figure}

\subsection{Ablation Study}
We sequentially add model components to validate the effectiveness of individual modules. Table \ref{table:Ablation} summarizes our results on the JRDB validation set. Here, we focus on the affinity matrices generated using transformer-based feature tokens. A single affinity matrix provides correspondences across two point cloud scans. First, we introduce only self-attention on the affinities. Here, for a given affinity matrix, self-attention seeks consensus from the surrounding correlations of the same affinity matrix and suppresses erroneous correspondences. On the JRDB validation set, incorporating self-attention on affinity matrices boosts the MOTA by $\sim$14\%. Secondly, we add cross-attention on affinities. In this case,  the attention mechanism estimates the relevance of surrounding correlations across two different affinity matrices. Thus, using correlation cues from the subsequent affinity matrix, the quality of the given affinity matrix is improved. Intuitively, cross-attention on affinities leverages temporal context to boost tracking performance. We see an improvement of $\sim$5\%, by introducing cross-attention on the affinities already refined by the self-attention mechanism (refer to the last row in Table \ref{table:Ablation}). Figure \ref{fig:Ablation_Tracking} illustrates the significance of attention-on-affinities on overall tracking performance.

\section{Conclusion}
We proposed a 3D point cloud-based joint detection and tracking model that eliminates the dependency on external detectors for object tracking. To overcome the challenges associated with sparsity and uneven density of 3D point cloud data, we introduced an attention mechanism and refined the affinity matrices that measure the correspondences across point cloud scans. Exhaustive experiments and ablations validated the effectiveness and generalizability of our approach in 3D MOT tasks. For future work, we will explore innovative techniques for reducing the computational requirements associated with attention-on-affinities, to allow us to incorporate a larger temporal context in the network. 

\clearpage

\bibliographystyle{IEEEtran}
\bibliography{IEEEabrv}

\end{document}